\documentclass[10pt,twoside,twocolumn]{article}

\usepackage{simpleConference}
\usepackage[a4paper,total={6in, 8in}, margin=1in]{geometry}

\usepackage{cite}
\usepackage{textcomp}
\usepackage[utf8]{inputenc} %
\usepackage[T1]{fontenc}    %
\PassOptionsToPackage{obeyspaces}{url}
\usepackage{hyperref}       %
\usepackage{booktabs}       %
\usepackage{amsfonts}       %
\usepackage{amsmath,amssymb,amsfonts,bm}    %
\usepackage{graphicx}
\usepackage{caption}
\usepackage{subcaption} %
\usepackage{stackengine}
\usepackage[algo2e, noend, ruled, vlined, english,linesnumbered]{algorithm2e}
\usepackage{transparent}
\usepackage{enumitem}

\usepackage{xspace}
\newcommand{\expnumber}[2]{{#1}\mathrm{e}{#2}}
\newcommand{\ie}{i.\nolinebreak[4]\hspace{0.125em}\nolinebreak[4]e.\@\xspace}
\newcommand{\eg}{e.\nolinebreak[4]\hspace{0.125em}\nolinebreak[4]g.\@\xspace}
\newcommand{\papertitle}{Input Selection for Bandwidth-Limited Neural Network Inference}

\newcommand{\x}{\bm{x}}
\newcommand{\maskedx}{\hat{\bm{x}}}
\newcommand{\y}{y}
\newcommand{\mask}{\bm{m}}
\newcommand{\maskf}{m}

\newcommand{\maskdiscrete}{\bm{m}^D}

\newcommand{\X}{X}
\newcommand{\Y}{Y}
\newcommand{\thumbx}{\bar{\x}}
\newcommand{\trainset}{T}
\newcommand{\model}{f}
\newcommand{\Reals}{\mathbb{R}}
\newcommand{\nruns}{n_{runs}}

\newcommand{\lambdainit}{\lambda_{init}}
\newcommand{\lambdafactor}{\lambda_{fac}}
\newcommand{\loss}{\mathcal{L}}
\newcommand{\lossmodel}{\mathcal{L}_{\model}}
\newcommand{\lossmask}{\mathcal{Q}}
\newcommand{\maskmodel}{g}

\newcommand{\npattern}{n}
\newcommand{\nepoch}{n_{\text{epoch}}}

\newcommand{\nbatches}{n_{\text{batch}}}
\newcommand{\iwidth}{w}
\newcommand{\iheight}{h}
\newcommand{\ichannel}{k}

\newcommand{\patience}{p}
\newcommand{\ntarget}{C} %

\newcommand{\remotedataset}{\texttt{remote}}
\newcommand{\galaxy}{\texttt{galaxy10}}
\newcommand{\cifar}{\texttt{cifar10}}
\newcommand{\mnist}{\texttt{mnist}}
\newcommand{\fashmnist}{\texttt{f-mnist}}
\newcommand{\svhn}{\texttt{svhn}}
\newcommand{\ships}{\texttt{ship}}

\usepackage{tikz}
\usetikzlibrary{shapes,arrows,shadows}
\usetikzlibrary{mindmap,trees}
\definecolor{mygreen}{rgb}{0.553,0.682,0.063}
\definecolor{myblue}{rgb}{0.0,0.208,0.376}
\definecolor{mygray}{rgb}{0.906,0.906,0.906}

\def\Checkmark{\tikz\fill[scale=0.4](0,.35) -- (.25,0) -- (1,.7) -- (.25,.15) -- cycle;} 
\usepackage{wrapfig}
\usepackage[export]{adjustbox}

\usepackage{xcolor}
\definecolor{light-gray}{gray}{0.75}

\author{Stefan Oehmcke%
\thanks{Department of Computer Science, University of Copenhagen}
\and Fabian Gieseke%
\thanks{
Department of Information Systems, University of Münster}\;\footnotemark[2]
}

\ifpdf
\hypersetup{
  pdftitle={\papertitle},
  pdfauthor={Stefan Oehmcke and Fabian Gieseke}
}
\fi

\newcommand\relatedversion{}
\renewcommand\relatedversion{\thanks{
The code is available at \protect\url{https://github.com/StefOe/selection-masks}}}

\begin{document}

\title{\papertitle\relatedversion}

\date{}

\maketitle

\begin{abstract} \small\baselineskip=9pt
Data are often accommodated on centralized storage servers. 
This is the case, for instance, in remote sensing and astronomy, where projects produce several petabytes of data every year. 
While machine learning models are often trained on relatively small subsets of the data, the inference phase typically requires transferring significant amounts of data between the servers and the clients. 
In many cases, the bandwidth available per user is limited, which then renders the data transfer to be one of the major bottlenecks.
In this work, we propose a framework that automatically selects the relevant parts of the input data for a given neural network.
The model as well as the associated selection masks are trained simultaneously such
that a good model performance is achieved while only a minimal amount of data is selected. 
During the inference phase, only those parts of the data have to be transferred between the server and the client.
We propose both instance-independent and instance-dependent selection masks. 
The former ones are the same for all instances to be transferred, whereas the latter ones allow for variable transfer sizes per instance.
Our experiments show that it is often possible to significantly reduce the amount of data needed to be transferred without affecting the model quality much.

\end{abstract}

\section{Introduction.}
The data volumes have increased dramatically in various domains. 
Often, centralized storage servers/clusters are used to accommodate the collected data, which are then accessed by many users. 
This is the case in remote sensing, where current projects produce petabytes of satellite data every year~\cite{Landsat,Sentinel}. 
The application of a machine learning model in this field to, \eg, monitor changes on a global scale or to search for objects, often requires ``scanning'' all the data and, thus, induces the transfer of large amounts of data between the server and the client that executes the model~\cite{Reichstein19}. 
Similar data-intensive scenarios can be found in other disciplines as well. 
For instance, astronomers also resort to centralized storage servers such as the ones for the Sloan Digital Sky Survey~\cite{SDSS2019}. 
The data transfer between such servers and (thousands of) clients is already restricted today (\eg, via a limited bandwidth per user) and will become a serious bottleneck for projects such as the Large Synoptic Sky Survey~\cite{LSST2019} or the Square Kilometre Array~\cite{SKA2017}, which will be fully operational within the next few years and which will yield petabytes of data every month.

While the reduction of the training and inference runtimes have received considerable attention~\cite{CoatesHWWCN13,Han2015,GordonENCWYC18,NIPS2016_6250,pmlr-v70-kumar17a,XuKWC2013}, comparatively little work has been done regarding the transfer of data induced by such server/client based scenarios.
In this work, we propose a framework that allows to automatically select those parts of the input data that are relevant for a given task and an associated model. 
In particular, we aim at scenarios, in which large amounts of data reside on a public storage server and where it is, in general, \emph{not} possible for the user to execute code on the server side.
\begin{figure}
    \centering
     \resizebox{.99\columnwidth}{!}{
      \input{figures/network_motivation_new.tex}
      }
    \caption{
End-to-end training of input selection masks and a task model to achieve both, a minimal amount of data selected by the masks as well as a good model performance.
During the inference phase, only the selected parts have to be transferred. 
\label{figure:pipeline}
}
\end{figure}
Our framework allows to learn masks that are adapted to the specific transfer capabilities offered by the server (\eg, if the server permits to select only certain channels or parts of the images), which can then be used to significantly reduce the amount of data needed to be transferred.\footnote{For instance, the \protect\href{https://developers.planet.com/docs/apis/}{\emph{Planet API}} allows to select input channels or to ``clip'' data before transmission.
} The masks as well as the model are optimized simultaneously in an end-to-end way to achieve both a minimal amount of data being selected by the masks and a good model performance, see Figure~\ref{figure:pipeline}.
During the inference phase, only the selected parts have to be transferred.
In addition to such ``static'' masks, we also consider scenarios where reduced versions of the data (\eg, thumbnails) are available on the server side that serve as basis for a dynamic selection of relevant input data for a given instance.
Our experiments show that both the static and dynamic selection masks can be used to significantly reduce the amount of data that must be transferred during the inference phase without sacrificing much of the model performance.

\section{Background.}

The transfer of data during the inference phase is an active field of research~\cite{NIPS2016_6250,lossyAE,xu2018scaling,lu2018deeppink,pmlr-v97-balin19a,lemhadri2021lassonet}.
Two different lines of approaches exist: (a) feature extraction, where the original features get modified, and (b) feature selection, where a subset of the original features is chosen.
In this work, we consider feature selection scenarios, where the user can select and slice the data, but is not able to perform any computations on the server side, which excludes the use of feature extraction schemes. 
For example, a neural network cannot be applied on the server.
Recently, unsupervised approaches based on autoencoders have been proposed that aim at selecting a pre-defined number of relevant input pixels. 
However, these methods loose the spatial information present in the data during the selection process~\cite{pmlr-v97-balin19a,han2018autoencoder}.
LassoNet~\cite{lemhadri2021lassonet} has been proposed as a supervised and unsupervised feature selection scheme.
It is, however, only applicable to fully-connected networks and not to the more prominent convolutional networks.

We conduct a gradient-driven search to find suitable weight assignments for the selection masks (defined below).
As detailed below, an exhaustive search for finding an optimal feature selection is computationally intractable.
An alternative to our approach are greedy schemes that, \eg, incrementally select input channels or pixels (\ie, similar to forward/backward feature selection methods).
However, these approaches also quickly become computationally infeasible in case many features (\eg, pixels or channels) are given. 
Another approach is based on iteratively selecting features according to a pre-calculated ranking, for instance based on the feature importance values induced by random forests~\cite{Breiman2001,geurts2006extremely} or based on a principal feature analysis~\cite{lu2007feature}. 
However, these methods neglect the feature structure, in particular spatial correlations, which is vital for many tasks. 
In addition, they usually yield sub-optimal results since they are not trained simultaneously with the task model in an end-to-end fashion.

Our framework is different from these works in the sense that we conduct supervised search with a more general and  flexible class of selection schemes (\eg, channel-, \mbox{block-,} or pixel-wise selection, see below).
Furthermore, instead of fixing the number of selected features beforehand, our approach iteratively reduces the number of features through gradient information updates, which provide more choices for the trade-off between accuracy and transfer cost.
There is also no restriction to specific network architectures (e.g., fully-connected networks), meaning any architecture can be used after the selection.
In addition to static scenarios, our approach allows dynamic selection of required input data per instance, which %
is a novel approach.

\section{Automatic Input Selection.}
The goal of our framework is to reduce the amount of data that needs to be transferred in order to apply a given neural network model for a specific task.
For the sake of simplicity, we focus on image data and classification in this work. 
Our approach can, however, also be applied to other types of data such as video, time series, or unstructured data as well as other types of tasks, such as regression or segmentation.

\subsection{Input Selection Masks.}

\begin{figure}
    \centering
    
     \begin{subfigure}[b]{0.3\linewidth}
       \resizebox{\linewidth}{!}{%
         \input{figures/select_special_any.tex}
       }
       \caption{\texttt{block}}
     \end{subfigure}%
    \hfill%
    \begin{subfigure}[b]{0.3\linewidth}
      \resizebox{\linewidth}{!}{%
        \input{figures/select_channel_any.tex}
      }
    \caption{\texttt{channel}}
    \end{subfigure}%
    \hfill%
    \begin{subfigure}[b]{0.3\linewidth}
      \resizebox{\linewidth}{!}{%
        \input{figures/select_pixel_any.tex}
      }
    \caption{\texttt{pixel}}
    \end{subfigure}%
    \caption{%
    Proposed masks to select data. 
    While the final masks are discrete, differentiable surrogates are used during the training process.
    }
\label{fig:selection_masks}
    \vspace{-5pt}
\end{figure}
Let \(\maskdiscrete\) be a \emph{selection mask} that allows to choose certain parts of an image~\mbox{\(\x \in \Reals^{\iwidth \times \iheight \times \ichannel}\)} with width~\(\iwidth\), height~\(\iheight\), and number of channels~\(\ichannel\), such as specific input channels or individual pixels, see Figure~\ref{fig:selection_masks}.
The generic selection scheme presented here is \texttt{block} selection, where the input data are divided into ``blocks''.
A partition in $b_\iwidth \times b_\iheight$ blocks is achieved via a mask of the form $\maskdiscrete \in \{0,1\}^{b_\iwidth \times b_\iheight \times k}$. %
As an example, a mask \(\maskdiscrete\) with~\(\maskdiscrete_{[2,1,3]}=1\) would correspond to selecting the second block in the first row of the third channel, whereas~\(\maskdiscrete_{[2,2,1]}=0\) would correspond to deselecting the second block in the second row of the first channel.
The case \(b_\iwidth = \iwidth\) and \(b_\iheight = \iheight\) yields \texttt{pixel} selection masks of the form~$\maskdiscrete \in \{0,1\}^{\iwidth \times \iheight \times k}$, which allow to select individual pixels per channel. 
Furthermore, the case \(b_\iwidth = 1\) and \(h_\iheight=1\) yields \texttt{channel} selection masks, where a mask \(\maskdiscrete \in \{0,1\}^{1 \times 1 \times k}\) allows to select specific channels of the image~\(\x\).

\subsection{Learning Static Masks.}
Training such a discrete mask that is well-suited for a given network and all the instances available for a specific task can be challenging.
In case a single mask is applied for all instances, we call the mask \emph{static} (since it remains the same regardless of the particular input image).
Let \( T = \{(\x_1, \y_1), \ldots, (\x_\npattern,\y_\npattern) \} \subset \X \times \Y\) be a training set consisting of images~\(\x_i \in \X= \mathbb{R}^{\iwidth\times \iheight\times\ichannel} \) with associated class labels \(\y_i \in \Y=\{1,\ldots,\ntarget\}\), where \(\ntarget\) is the number of classes.
Masking is implemented by element-wise multiplication \(\maskdiscrete \odot \x \) of the mask~\(\maskdiscrete\) with a given input image \(\x\), which sets deselected blocks to zero.
In order to apply this step, we need to ensure that if the input image \(\x\) and the corresponding mask \(\maskdiscrete\) have different shapes (\ie, \(b_\iwidth \neq \iwidth\) or \(b_\iheight \neq \iheight\)), the first two axes are broadcasted (via nearest neighbor interpolation), so it always yields a mask \(\maskdiscrete \in \{0,1\}^{\iwidth \times \iheight \times \ichannel}\) with the same shape as the input.

The goal of the training process is to find suitable weight assignments for both, the selection mask~\(\maskdiscrete\) and the neural network~\(\model : \X \rightarrow \Y\) that is being considered for the task at hand.
Simultaneous training of mask and network is achieved by minimizing the loss function
\begin{equation}
    \loss_{f,\maskdiscrete} \left(\hat{\y}, \y \right) = \lossmodel\left(\hat{\y}, \y\right) + \lambda \cdot \lossmask\left(\maskdiscrete\right),
\end{equation}
where \(\hat{\y}=\model(\maskdiscrete \odot \x)\) is the prediction for a given image \(\x\) with associated class label \(\y\), \(\lossmodel\) a user-defined task/model %
loss function (\eg, cross-entropy), and \(\lossmask\) a \emph{mask loss} that penalizes selections made by the mask \(\maskdiscrete\).
For the mask loss, various choices are possible. 
In this work, we consider the standardized L1-loss:
\begin{equation}\label{alg:maskloss}
    \lossmask\left(\maskdiscrete\right) = 
    \frac{\sum_{i=1}^{b_\iwidth}\sum_{j=1}^{b_\iheight}\sum_{l=1}^{\ichannel} \maskdiscrete_{[i, j, l]}}{b_\iwidth b_\iheight \ichannel} 
\end{equation}
It's constant gradient helps to gradually deactivate mask entries.
The parameter $\lambda \in \mathbb{R}^+$ increases the weight of $\lossmask$, which we adapt heuristically to overcome stagnation (see Section~\ref{sec:init-params-post}).
This mask loss is a direct representation of how much data is selected and needs to be transferred (\eg, 0.25 is 25\% of the data).

\subsubsection{Optimizing Discrete Masks.} \label{sec:opt_discrete}
Naturally, search schemes that aim at finding optimal discrete weight assignments for the mask w.r.t.~\(\loss_{f,\maskdiscrete}\) by testing all possible assignments are computationally infeasible.
Simple greedy approaches such as forward/backward selection of channels become computationally very demanding and are, thus, generally ill-suited (especially for pixel- or block-wise selection).
While adapting the selection mask~\(\maskdiscrete\) directly via gradient descent is, in general, possible, doing so yields imprecise gradient information due to its binary value space.

We therefore introduce a mask model~\(\maskf: \X \rightarrow \X\) %
with a trainable weight matrix~\(\mask \in \mathbb{R}^{b_\iwidth\times b_\iheight\times\ichannel}\) that outputs the selected input~\(\maskedx\). 
In the static case considered so far, the associated weights of the model \(\maskf\) are independent of the particular (image) instance; for the dynamic case, the weights will depend on the particular instances, see Section~\ref{subsec:learning_dynmasks}. 
In both cases, the model output is given by
\begin{equation}\label{eq:discrete_fwd}
    \maskf(\x) = \maskdiscrete \odot \x = \lfloor \sigma\left(\mask\right) \rceil \odot \x
\end{equation}
during the forward pass of the training process, where the rounding operator $\lfloor\cdot\rceil$ discretizes the weights (e.g., $\lfloor0.2\rceil = 0$ and $\lfloor0.8\rceil = 1$) and where $\sigma:\Reals \rightarrow \Reals$ is the sigmoid function with $\sigma(z) = \frac{1}{1 + \expnumber{}{}^{{-z}}}$ that is applied in an element-wise fashion to the weight vector $\mask$.
For the backward pass, a real-valued surrogate is used to obtain a continuous local gradient for the model $m$ (that depends on the trainable weight vector $\mask$). %
This surrogate ignores the rounding operator (since it is not differentiable) and only resorts to the derivative of the sigmoid function, i.e., 
\begin{equation}\label{eq:discrete_bwd}
    \frac{d~\maskf}{d~m_j} = %
    \frac{\expnumber{}{}^{-m_j}}{(1 + \expnumber{}{}^{-m_j})^2} x_j,
\end{equation}
for the individual weights $m_j$ of the weight vector $\mask$.
This enables effective backpropagation since the sigmoid function is a differentiable function that can be used as surrogate for the rounding operator.\footnote{More precisely, one can consider $\bar{\sigma}(z) = \sigma(\frac{z}{\tau})$, which approximates the rounding operator $\lfloor\cdot\rceil$ for $\tau \rightarrow 0$. 
This approximation of the non-differentiable function is similar to the one used in~\cite{gumbelTrick2014}; no random noise has to be added in this case though.
There are also parallels to straight through estimators used in model pruning and compression~\cite{binNetworks,discreteModelComp,exploreModelComp}, but instead of using an identity function and clipping the gradient to be between 0 and 1, we utilize Equation~\ref{eq:discrete_bwd}. %
For all experiments reported in this work, we used $\tau=1$, i.e., a normal sigmoid function was used as approximation.}

\subsubsection{Training.}

\begin{algorithm2e}[t]

\DontPrintSemicolon
\SetKwInOut{Input}{Input}
\caption{\texttt{LearnMasks}(\(\model, \maskf, \trainset\))}
\label{alg:learnselectionmasks}
\Input{neural network \(\model\), mask model \(\maskf\), and training set \(\trainset\)}
\(\mask \leftarrow \) \texttt{InitAllMasks()} \label{line:init_masks}\\ %
\(\lambda \leftarrow \) \texttt{InitLambda()} \label{line:init_lambda_tau}\\
    \For{\(i \leftarrow 1\) \KwTo \(\nepoch\) \label{line:nepoch}}{
        \For{\(j\leftarrow 1\) \KwTo \(\nbatches\)\label{line:nbatches}}{
            \(\x, \y \leftarrow \texttt{GetBatch}(\trainset) \)
            \label{line:load}\\
            \(\maskedx \leftarrow \maskf(\x)\) \label{line:maskx} \\ %
            \(\hat{\y} \leftarrow \model\left(\maskedx\right) \)\ \label{line:pred}\\
            \(\loss_{f,\maskdiscrete} \leftarrow \lossmodel(\hat{\y}, \y) + \lambda \lossmask \left(\maskdiscrete\right)\) \label{line:loss}\\
            \(\model, \maskf \leftarrow\texttt{Optimize}\left(\model, \maskf, \loss_{f,\maskdiscrete}\right)\) \label{line:optimize}\\
        }

    \(\lambda \leftarrow \) \texttt{AdaptLambda\(\left(\lambda, \lossmask \left(\maskdiscrete\right)\right)\)} \label{line:adapt_lambda_tau} \\
    \texttt{ModelCheckpoint\(\left(\model, \maskf, \lossmask \left(\maskdiscrete\right)\right)\)}\label{line:model_checkpoint}\\
    }

\end{algorithm2e}

Our training procedure for learning a suitable mask and network model is given by~\texttt{LearnMasks} in Algorithm~\ref{alg:learnselectionmasks}:
The weights associated with the mask model as well as the trade-off parameter $\lambda$ are initialized in Line~\ref{line:init_masks} and~\ref{line:init_lambda_tau}, respectively. 
Both the mask model~\(\maskf\) and the network~\(\model\) are trained simultaneously by iterating over a pre-defined number~\(\nepoch\) of epochs,
each being split into \(\nbatches\)~mini-batches (for the sake of clarity, we consider a batch size of 1).
For each batch, a discrete mask~\(\maskdiscrete\) is computed and element-wisely multiplied (\(\odot\)) with the input~\(\x\) via the mask model~\(\maskf\) to return the masked image~\(\hat{\x}\), see again Equation~\eqref{eq:discrete_fwd}. 
The induced prediction~\(\hat{y}\) is then used to compute the task loss~\(\lossmodel(\hat{\y}, \y)\). 
Both, the task loss~\(\lossmodel(\hat{\y}, \y)\) and the mask loss~\(\lossmask (\maskdiscrete)\) are optimized simultaneously via the procedure \texttt{Optimize} in Line~\ref{line:optimize} using standard optimizers such as stochastic gradient descent or Adam~\cite{adam}. 
The influence of the mask loss is gradually increased by adapting~\(\lambda\) after each epoch via the procedure \texttt{AdaptLambda} (see Section~\ref{sec:init-params-post}). 
Throughout the overall process, the procedure \texttt{ModelCheckpoint} assesses the mask losses of the intermediate models and stores them according to user-defined criteria.

\subsubsection{Initialization, Parameters, and Post-Training.} \label{sec:init-params-post}
We initialize the weight matrix~\(\mask\) of the mask model~\(\maskf\) via the \texttt{InitAllMasks} procedure.
Note that the decision boundary for the discrete mask is \(0.5\) due to the rounding operation applied in Equation~\eqref{eq:discrete_fwd} for discretization.
Therefore, we initialize the weight matrix~\(\mask\) with values larger than \(0\) since \(\sigma(0)=0.5\).
This ensures that the mask model~\(\maskf\) initially retains all blocks, thereby allowing the prediction model to operate on the complete data at the beginning. 
This, in turn, ensures that the prediction model provides useful gradient information about the less important parts of the input.
If parts of the input are removed too fast, the prediction model may degrade too quickly. %
Thus, we initialize the entries of the weight matrix~\(\mask\) with random values drawn from a normal distribution \(\mathcal{N}(\mu, \sigma) \), where \(\mu \) is a constant larger than 1 and \(\sigma\) a small positive value (\(\mu=2\) and \(\sigma=0.01\) in all our experiments).

The procedure \texttt{InitLambda} initializes $\lambda$,
which determines the trade-off between the task loss $\lossmodel$ and the mask loss $\lossmask$.
Initially, $\lambda$ is set to a small value (\eg, $\lambda=0.1$) to ensure that the input data is not directly removed at the beginning of the training process.
The influence of $\lambda$ is then gradually increased until $\nepoch$ epochs have been processed.
Since the range of possible values for the model loss~\(\lossmodel\) is generally unknown a priori, we resort to a scheduler that increases~\(\lambda\) through \texttt{AdaptLambda} in Line~\ref{line:loss} of Algorithm~\ref{alg:learnselectionmasks} in case the overall error $\loss_{f,\maskdiscrete}=\lossmodel + \lambda \lossmask$ has not decreased for a specific amount $\patience$ of epochs. 
Thus, the $\lambda$-scheduler behaves similarly to standard learning schedulers. 
However, instead of decreasing the learning rate, the value for \(\lambda\) is increased by a user-defined factor $\lambdafactor$ (\eg, $\lambdafactor=1.1$). 

During training, the intermediate models and masks are stored via the procedure \texttt{ModelCheckpoint} and the best-performing combination can be selected in the end.
It is also possible to store the best-performing models for each interval of a user-defined partition of the amount of data to be removed (\eg, (5\%, 10\%, ..., 100\%)) to account for varying bandwidth availability.
In general, the performance of the models can also be slightly improved by continuing training the weights of the model $\model$ for several epochs without changing the mask weights anymore (post training).

\subsection{Learning Dynamic Masks.}
\label{subsec:learning_dynmasks}

The approach outlined above is used to obtain task-specific input selection masks that are the same for all instances.
In addition to such ``static'' masks, we introduce scenarios in which one additionally has access to a thumbnail/reduced version $\thumbx$ for each input image $\x$ (see Section~\ref{sec:experiments}). %
As described next, we use these reduced versions to obtain instance-based selection masks with a small constant overhead.
Compared to the static masks, these dynamic masks yield further reductions in data transfer.

\begin{figure}
    \centering
     \resizebox{\linewidth}{!}{
      \includegraphics{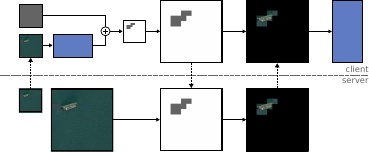}
      }
    \caption{The thumbnail model $\maskmodel$ (left blue rectangle) identifies the relevant parts of input data based on a thumbnail~$\thumbx$. Only those parts of the image $\x$ have to be transferred and are processed by the task model $\model$ (right blue rectangle).\label{fig:dynmasks}}
\end{figure}
Figure~\ref{fig:dynmasks} outlines the dynamic mask approach:
The basic idea is to decide which parts of a given instance~$\x$ need to be transferred based on the thumbnail~$\thumbx$. 
To that end, the instance-based masks are generated via a separate thumbnail model~\(\maskmodel :\overline{\X} \rightarrow \mathbb{R}^{b_\iwidth\times b_\iheight\times\ichannel}\) that receives a thumbnail~\(\thumbx\) and outputs instance-dependent mask weights (\eg, \(\maskmodel\) could be a small convolutional neural network).
Effectively, we replace the weight matrix~\(\mask\) of the mask model \(\maskf\) by the output of the thumbnail model~\(\maskmodel\) in \eqref{eq:discrete_fwd} and \eqref{eq:discrete_bwd}, \ie:
\begin{equation}
    \mask = \maskmodel(\thumbx) + c
\end{equation}
Here, \(c\) is a positive constant (\eg, \(c=1\)). 
Further, the weights of the thumbnail model $\maskmodel$ are initialized in such a way that  $\maskmodel(\thumbx) \approx \mathbf{0}$ at the beginning.
This ensures that the masks select all data initially.
These instance-based masks have the same size as the thumbnails and are, hence, expanded (\eg, via nearest neighbors interpolation) to  the dimensions of the original input $\x$. %

As for the static case, both the original model~$\model$ and the thumbnail model \(\maskmodel\) 
are trained simultaneously (on the client). 
The discretization approach also remains the same.
Note that the thumbnail model~$\maskmodel$ has to conduct a conceptually simpler task than the task model~$\model$: It only has to identify those parts of the input data~$\x$ that are potentially relevant for $\model$; it does not have to address the final learning task. 

During the inference phase, the thumbnail, the instance-dependent mask, and the selected data must be transferred, which creates a small constant overhead for each instance.
However, depending on the complexity of the instance, a dynamically selected input image may require significantly fewer selections than a static mask, which must be well suited for all possible input images.
This higher reduction can often outweigh the price for the small constant overhead.
In addition, each input image is selected by masks whose block size is smaller than the original input, which allows efficient block-wise encoding of the data to be transmitted.
As shown in Section~\ref{exp:dynamic}, such dynamic selections significantly reduce data transfer costs compared to a static selection.

\begin{table}
    \caption{
    Datasets and Models\label{tab:datasets}
    }
    {
    \centering
    \setlength\tabcolsep{1.5pt}
    \resizebox{\linewidth}{!}{
    \begin{tabular}{lrrrrrrlr}
    \toprule
    Dataset       & \#train & \#hold-out & \#class & \(\iwidth\)   & \(\iheight\)  & \(c\) & model & lr\\ 
    \midrule
    \remotedataset& {24694}&  {24694} & 12 & {35}& {35}& {36} & AllConvNet & \(\expnumber{1}{-3}\) \\
    \galaxy & {19606}&  {2179} & 10 & {69}& {69}& {3} & ResNet20 & \(\expnumber{1}{-4}\) \\
    \cifar        & {50000} & {10000} & 10 &{32}& {32}& {3} & ResNet20 & \(\expnumber{1}{-5}\) \\
    \mnist        & {60000} & {10000} & 10 &{28}& {28}& {1} & LeNet5 & \(\expnumber{1}{-3}\) \\
    \fashmnist        & {60000} & {10000} & 10 &{28}& {28}& {1} & LeNet5 & \(\expnumber{1}{-3}\) \\
    \svhn         & {73257} & {26032} & 10 &{32}& {32}& {3} & ResNet20 & \(\expnumber{1}{-5}\) \\
    \ships         & {175548} & {17008} & 2 &{768}& {768}& {3} & Squeezenet & \(\expnumber{1}{-5}\) \\
     \bottomrule
    \end{tabular}
    }
    }
    \vspace{-10pt}
\end{table}

\begin{figure*}
\centering
\begin{subfigure}[b]{0.32\linewidth}
   \centering
   \resizebox{\linewidth}{!}{
            \input{figures/experiments/channel_remote.tex}
        }
    \caption{\texttt{channel} on \remotedataset \label{fig:exp_channels}}
\end{subfigure}%
\begin{subfigure}[b]{0.32\linewidth}
   \centering
   \resizebox{\linewidth}{!}{
      \input{figures/experiments/quad_svhn.tex}
      }
  \caption{(\(12\times 12\))-\texttt{block} on \svhn \label{fig:exp_quad}}
\end{subfigure}%
\begin{subfigure}[b]{0.32\linewidth}
   \centering
   \resizebox{\linewidth}{!}{
      \input{figures/experiments/pixel_galaxy10.tex}
      }
    \caption{\texttt{pixel} on \galaxy \label{fig:exp_pixels}}
\end{subfigure}
  \caption{
The black line is the average value of the runs and the individual runs are displayed in different colors. Note that the fluctuations are induced by the training process during which the input data are partially ignored.
Performance at $\nepoch= 0$ represents performance of the pre-trained network.
  }
\end{figure*}
\begin{figure}[!t]
\vspace{5pt}
\centering
\begin{subfigure}[b]{\linewidth}
\centering
\resizebox{.9\linewidth}{!}{
\includegraphics{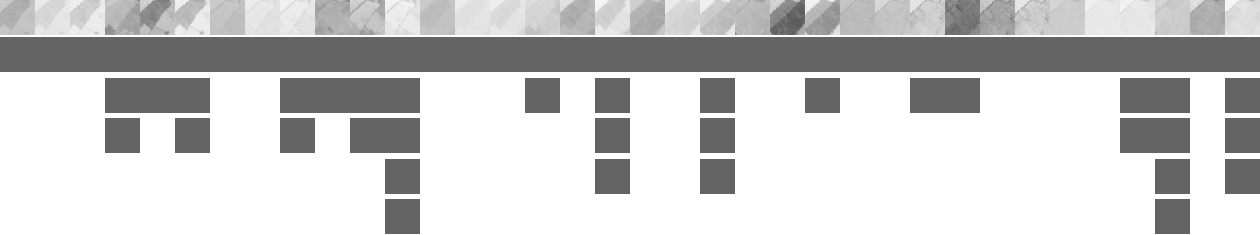}
}
\caption{\texttt{Channel} selection on \remotedataset}
\label{fig:channel_mask}
\end{subfigure}

\begin{subfigure}[b]{\linewidth}
\centering
    \resizebox{.9\linewidth}{!}{%
    \includegraphics{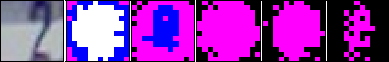}}%
    \caption{(\(12\times 12\))-\texttt{block} selection on \svhn}%
    \label{fig:example_svhn_blockwise}
\end{subfigure}
\begin{subfigure}[b]{\linewidth}
\centering
    \resizebox{.9\linewidth}{!}{%
    \includegraphics{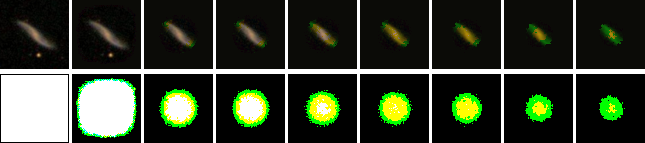}}%
    \caption{\texttt{pixel} selection on \galaxy}%
    \label{fig:block-example_supernova_pixelwise}
\end{subfigure}
\vspace{-14pt}
\caption{%
In Figure~(a), the selection process is sketched, whereby each row represents a different epoch (from top to bottom: {example instance}, {0}, {50}, {100}, {150}, {200}) and each column represents a channel.
For Figure~(b), the example image is provided (left) along with the mask development during the training process. Figure~(c) shows the progression of the image instance (i.e. the selected pixels of the image; at the end, central pixels from the green channel are selected).}
\end{figure}

\section{Experiments.}
\label{sec:experiments}

We considered several classification datasets and network architectures, see Table~\ref{tab:datasets}. 
In addition to the well-known \cifar, \mnist, fashion-\mnist\ (\fashmnist), and \svhn\ datasets \cite{cifar10,lecun2010mnist,fmnist,37648}, we resorted to three more datasets from remote sensing and astronomy, respectively:
For each instance of \remotedataset, one is given an image with 36 channels originating from six multi-spectral satellite image bands available for six different dates~\cite{PRISHCHEPOV2012195}.
The learning goal of \remotedataset\ is to predict the type of change occurring in the central pixel of each image.
The dataset \protect\href{https://astronn.readthedocs.io/en/latest/galaxy10.html}{\galaxy}\ is dedicated to detecting different types of galaxies based on RGB images from the Sloan Digital Sky Survey~\cite{aihara2011eighth} with labels from Galaxy Zoo~\cite{lintott2008galaxy}.
Both \remotedataset\ and \galaxy\ are typical datasets for their respective domain, with the target objects being located in the centers of the images.
Finally, we considered the \ships\ dataset of the \protect\href{https://www.kaggle.com/c/airbus-ship-detection}{Airbus Ship Detection Challenge}.
We simplified the task from segmentation to classification (\ie, the task was to detect if a ship is visible in the image or not), halved the sizes of the original images (to obtain images of size $384 \times 384$), and used undersampling to balance the classes.
In contrast to the other datasets, the relevant information is not necessarily in the middle of an image, which renders static selection approaches less useful.

For all experiments, we considered a fixed amount of epochs and monitored the classification accuracy and mask loss~\(\lossmask\) on the hold-out set, see Table~\ref{tab:datasets}.
The mask loss is a direct measure of how much raw information needs to be transferred.
Depending on the application, further reductions might be achieved by using compression algorithms after the input selection.
Each experiment was conducted $\nruns=10$ times. %
We used pre-trained networks and optimizer parameters.

Our implementation is in PyTorch (version 1.5).
Except for the trade-off parameter~$\lambda$ and the learning rate for the mask model~\(\maskf\), all parameters were fixed (e.g. batch size of 128). 
The Adam~\cite{adam} optimizer with AMSGrad~\cite{amsgrad} and specific learning rates per model and dataset were employed, see Table~\ref{tab:datasets}.
More implementation details are available in the appendix.

\subsection{Static Selection Masks.}

We start by demonstrating the basic functionality of the different types of selection masks, an evaluation of the parameter~\(\lambda\), and a comparison with other static selection approaches.

\subsubsection{Selection Schemes.}\label{exp:basic_selection_schemes}

\begin{figure}
\begin{center}
\centering
\vspace{5pt}
\resizebox{.95\linewidth}{!}{
\input{figures/experiments/lambda_svhn.tex}
}
\end{center}
    \vspace{-.5cm}
\caption{Influence of \(\lambda\)\label{fig:lambda_test}}
\end{figure}

We first evaluated the behaviour of the \texttt{channel} selection scheme ($b_\iwidth=1$ and $b_\iheight=1$) on \remotedataset\ to select the most relevant of the 36 input channels.
Figure~\ref{fig:exp_channels} shows the outcome of the iterative selection process induced by Algorithm~\ref{alg:learnselectionmasks}.
Only if less than {25}\% of the channels were selected, the accuracy started to drop.
The evolution of this mask selection can be seen in Figure~\ref{fig:channel_mask}.

\begin{figure*}
{\raggedleft
\resizebox{0.315\linewidth}{!}{
\input{figures/experiments/dyn_mnist.tex}
}}\hfill{\centering
\resizebox{0.315\linewidth}{!}{
\input{figures/experiments/dyn_fashion.tex}
}}\hfill{\raggedright
\resizebox{0.315\linewidth}{!}{
\input{figures/experiments/random_cifar.tex}
}}
\begin{subfigure}[b]{0.315\linewidth}
\centering
\hspace{-.1cm}\stackunder[2pt]{\resizebox{0.2\linewidth}{!}{
\includegraphics{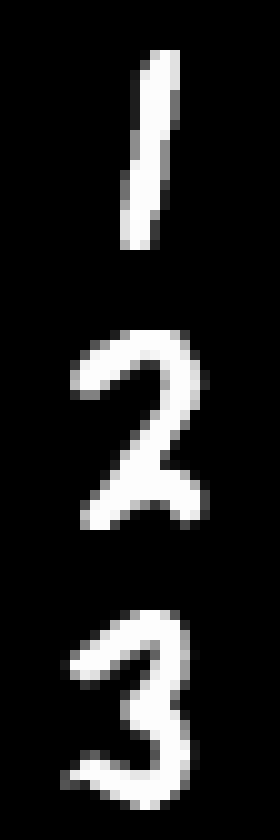}
}}{\small orig.}%
\centering
\hfill\stackunder[2pt]{\resizebox{0.2\linewidth}{!}{
\includegraphics{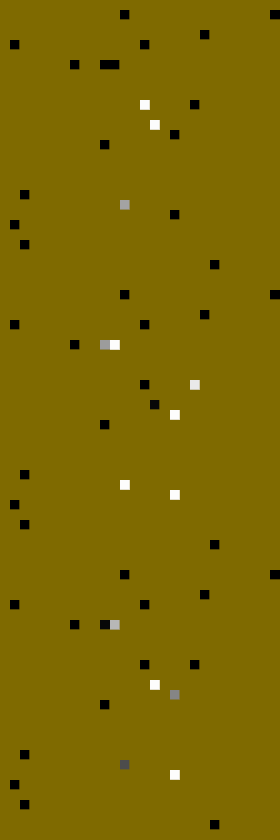}
}}{\small rand.}%
\centering%
\hfill\stackunder[2pt]{\resizebox{0.2\linewidth}{!}{
\includegraphics{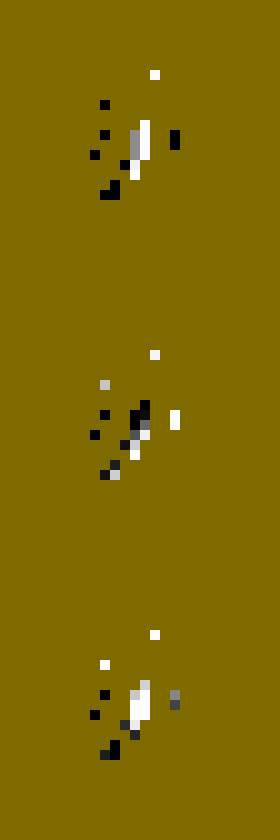}
}}{\small RF}%
\centering
\hfill\stackunder[2pt]{\resizebox{0.2\linewidth}{!}{
\includegraphics{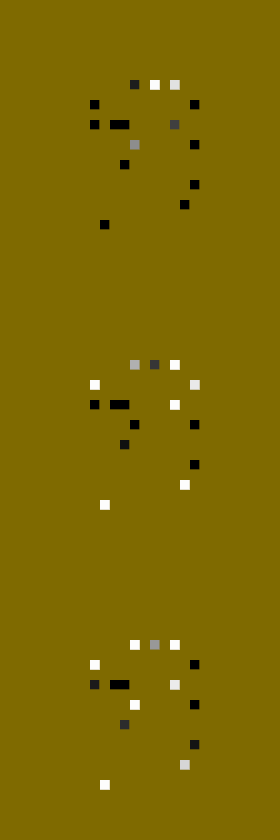}
}}{\small static}%
\centering
\hfill\stackunder[2pt]{\resizebox{0.2\linewidth}{!}{
\includegraphics{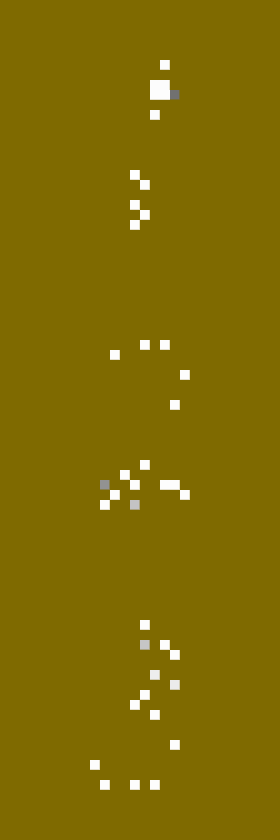}
}}{\small dyn.}%
\caption{\mnist}
\end{subfigure}%
\hfill\begin{subfigure}[b]{0.315\linewidth}
\centering
\stackunder[2pt]{\resizebox{0.2\linewidth}{!}{
\includegraphics{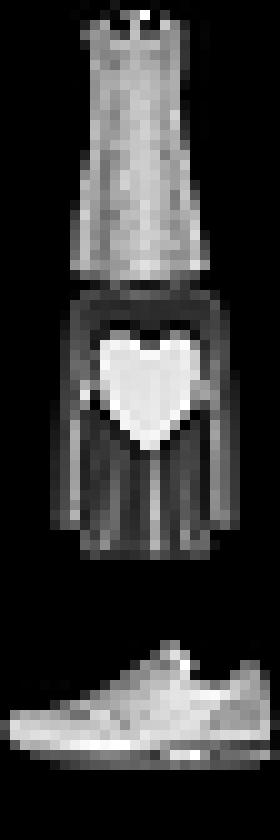}
}}{\small orig.}%
\centering
\hfill\stackunder[2pt]{\resizebox{0.2\linewidth}{!}{
\includegraphics{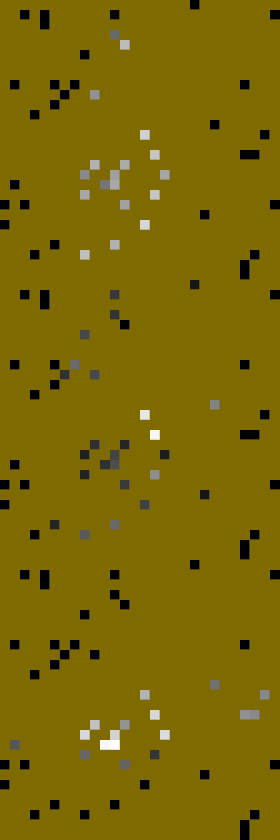}
}}{\small  rand.}%
\centering
\hfill\stackunder[2pt]{\resizebox{0.2\linewidth}{!}{
\includegraphics{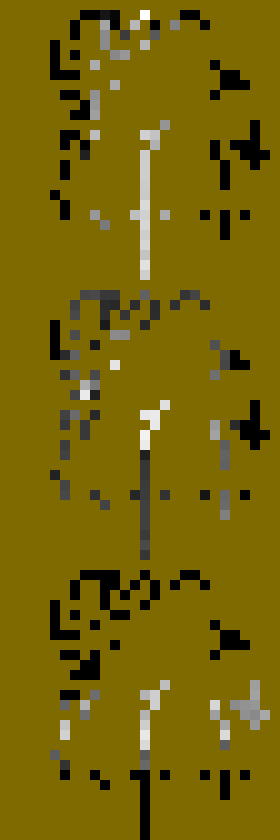}
}}{\small RF}%
\centering
\hfill\stackunder[2pt]{\resizebox{0.2\linewidth}{!}{
\includegraphics{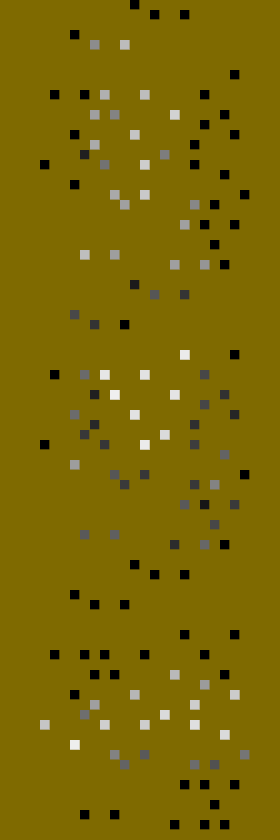}
}}{\small static}%
\centering
\hfill\stackunder[2pt]{\resizebox{0.2\linewidth}{!}{
\includegraphics{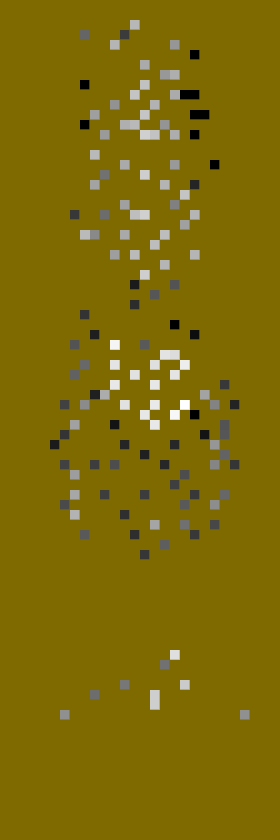}
}}{\small dyn.}%
\caption{\fashmnist}
\end{subfigure}%
\hfill\begin{subfigure}[b]{0.315\linewidth}
\centering
\stackunder[2pt]{\resizebox{0.2\linewidth}{!}{
\includegraphics{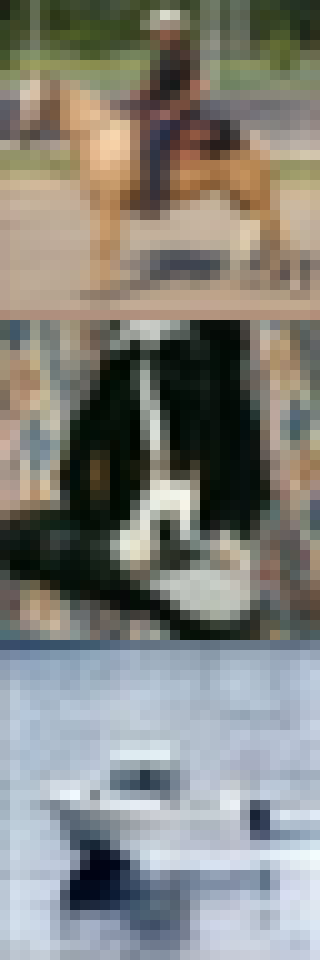}
}}{\small orig.}%
\centering
\hfill\stackunder[2pt]{\resizebox{0.2\linewidth}{!}{
\includegraphics{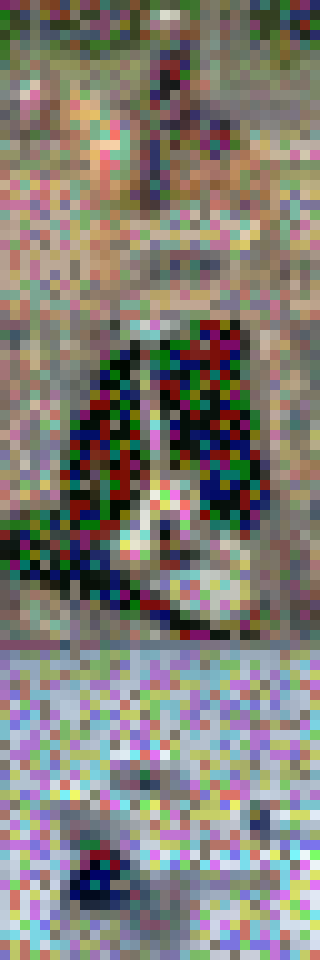}
}}{\small  rand.}%
\centering
\hfill\stackunder[2pt]{\resizebox{0.2\linewidth}{!}{
\includegraphics{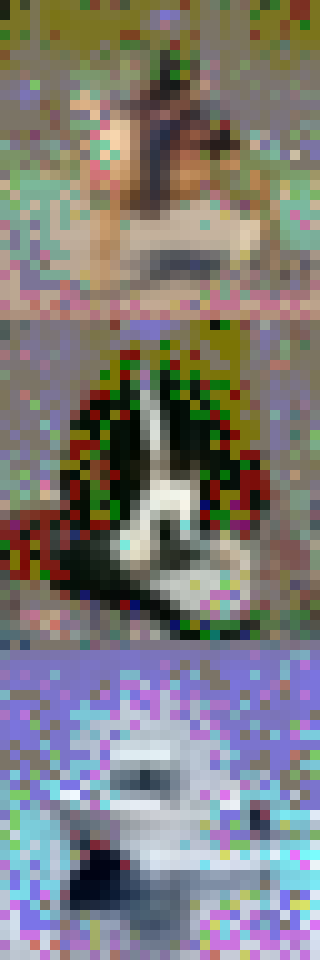}
}}{\small RF}%
\centering
\hfill\stackunder[2pt]{\resizebox{0.2\linewidth}{!}{
\includegraphics{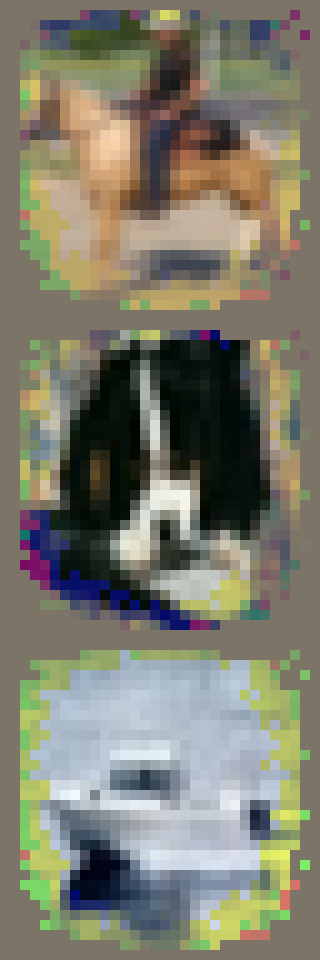}
}}{\small static}%
\centering
\hfill\stackunder[2pt]{\resizebox{0.2\linewidth}{!}{
\includegraphics{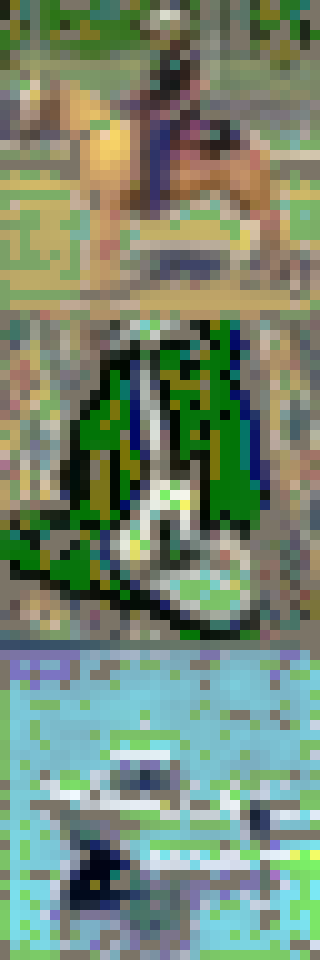}
}}{\small dyn.}%
\caption{\cifar}
\end{subfigure}
  \caption{
The top row shows the behaviour w.r.t.\ the mask loss \(\lossmask\). The bottom row shows instances for different \(\lossmask\): \mnist\ (left) \(\sim 2.42\%\), \fashmnist\ (middle) \(\sim 6\%\), and \cifar\ (right) \(\sim 66\%\).
Dark yellow color filling corresponds to removed pixels for \mnist\ and \fashmnist.
For \cifar\ grey indicates completely removed pixel while tints indicate that channels for a pixel are disabled (e.g., green tinted pixel indicates missing red and blue).
\label{fig:dyn_example}}
\end{figure*}

The \texttt{block} selection experiment was conducted on \svhn. %
The considered mask selected \(12 \times 12\) blocks ($b_\iwidth=12$ and $b_\iheight=12$) from each image per channel, see again Figure~\ref{fig:selection_masks}.
Figure~\ref{fig:example_svhn_blockwise} shows an instance and the progress w.r.t.\ $\nepoch$ for the RGB channels (purple: red and blue; yellow: green and red; white: all selected).
Figure~\ref{fig:exp_quad} shows the outcome of the iterative selection: Smooth optimization and a noticeable loss in accuracy around \(\lossmask=25\%\).
To investigate the behaviour of \texttt{pixel}-wise selection, we considered the \galaxy\ dataset.
The mask loss~\(\lossmask\) here reflects the summation over the selected pixels, where each pixel had a weight of \(\frac{1}{\iwidth \times \iheight \times \ichannel}\). 
The results of the iterative selection are provided in Figure~\ref{fig:exp_pixels}, whereas 
Figure~\ref{fig:block-example_supernova_pixelwise} shows an instance and the development of the masks w.r.t.~$\nepoch$.

\subsubsection{Influence of \texorpdfstring{\(\lambda\)}{lambda}.}
The initial assignment \(\lambdainit\) for \(\lambda\) as well as the factor \(\lambdafactor\) generally have a significant impact on the outcome.
Figure~\ref{fig:lambda_test} shows the influence of four different configurations given the \svhn\ dataset. %
A large \(\lambdainit\)~(red and green line) led to the mask loss \(\lossmask\) quickly decreasing, but it also caused a lower accuracy compared to the smaller \(\lambdainit\) values.
A smaller initial value for \(\lambda\) generally led to the selection process taking less input data away at the beginning. %
Accordingly, a large \(\lambdafactor\) led to a faster decrease w.r.t.\ \(\lossmask\).

\subsubsection{Comparison with Baselines.}

We compared our static selection approach with two direct competitors: random (backward) selection and the selection w.r.t.\ the feature importance values of a random forest (RF)~\cite{Breiman2001}.
In particular, we compared the approaches in the context of pixel-wise selections on \cifar, \mnist, and \fashmnist. %
To simulate a fair training process, we first created a ranking for each pixel based on randomness and the feature (\ie, pixel) importance values induced by an extra trees classifier~\cite{geurts2006extremely} with 100 trees, respectively.
Given the same task model $\model$ and the same amount of epochs, we then iteratively removed pixels to obtain the same loss \(\lossmask\) as our static approach.

Figure~\ref{fig:dyn_example} shows the results (the runs appear to be ``clustered'' since we saved the best accuracy in steps of 0.05-\(\lossmask\) intervals):
In can be seen that our static mask selection quickly diverges in accuracy from the random selection with decreasing \(\lossmask\).
The RF-based selection performed better than the random selection approach, but was outperformed by our approach, which shows the benefits of optimizing both the mask weights and the model weights simultaneously.
We also compared our results with the concrete autoencoder~(CAE)\cite{pmlr-v97-balin19a} approach, which 
selects a fixed (arbitrary) amount of pixels without retaining spatial structure and LassoNet~\cite{lemhadri2021lassonet}, which iteratively removes features but can only be applied to fully-connected networks. 
The CAE approach achieved a self-reported accuracy of about \(90.6\%\) on \mnist\ with a mask loss \(\lossmask\) of \(0.064\) (50 out of 784 pixels) and the LassoNet paper reports \(87.3\%\) at the same mask loss.
In comparison, our static selection approach yielded an accuracy of \(96.11\%\) given a lower mask loss \(\lossmask=0.047\). 
Similarly, on \fashmnist, the CAE achieved a self-reported accuracy of \(67.7\%\) given a mask loss \(\lossmask\) of \(0.064\), while our static mask yielded an accuracy of 83.86\% given a mask loss of \(\lossmask=0.045\).
It is worth stressing that the CAE and LassoNet approach resort to either random forest or fully-connected networks applied to the selected pixels, i.e., they cannot make use of predictors with spatial awareness.

\begin{figure*}
\begin{center}
\begin{subfigure}[b]{1\linewidth}
\qquad\quad\stackon[2pt]
{\stackon[2pt]{
\hspace{-.45cm}\resizebox{0.0925\linewidth}{!}{\includegraphics{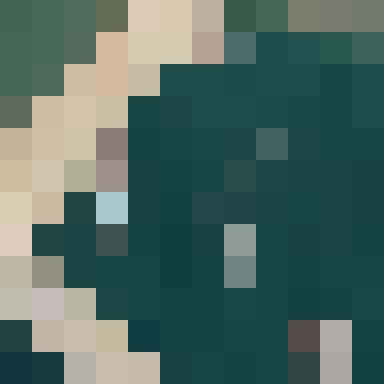}}~%
\resizebox{0.0925\linewidth}{!}{\includegraphics{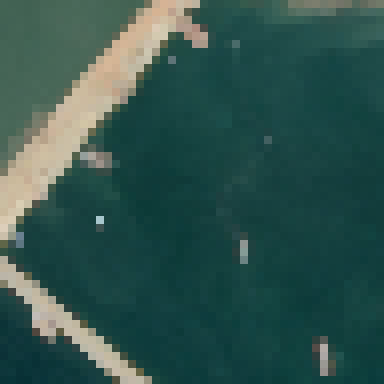}}
}{\stackon[2pt]{\small \(12\times 12\) \qquad \(48\times 48\)}{\small thumbnails}}\hspace{1cm}%
\resizebox{0.2\linewidth}{!}{\includegraphics{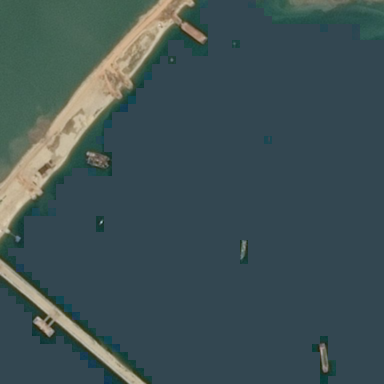}}\hspace{.2cm}%
\stackon[2pt]{\resizebox{0.2\linewidth}{!}{\includegraphics{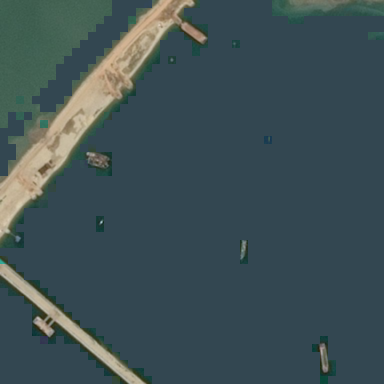}}}{\small \(48\times 48\)}\hspace{.2cm}%
\resizebox{0.2\linewidth}{!}{\includegraphics{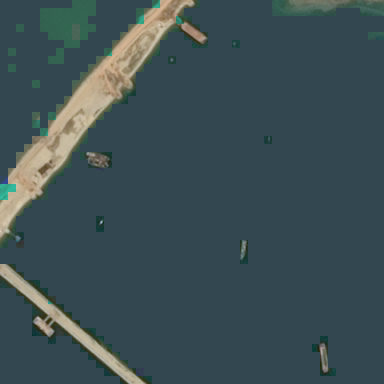}}%
}
{\stackon[2pt]{\resizebox{0.2\linewidth}{!}{
    \includegraphics{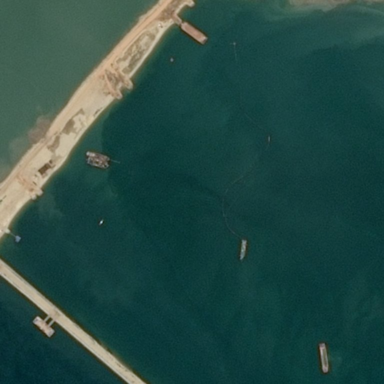}
}}{\small original}\hspace{1cm}%
\resizebox{0.2\linewidth}{!}{\includegraphics{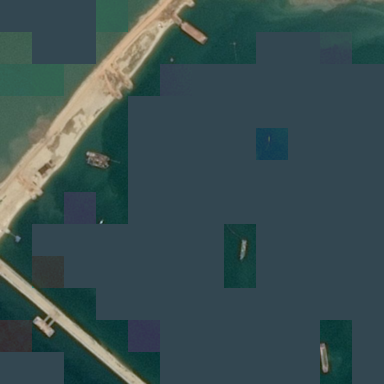}}\hspace{.1cm}%
\stackon[2pt]{
    \resizebox{0.2\linewidth}{!}{\includegraphics{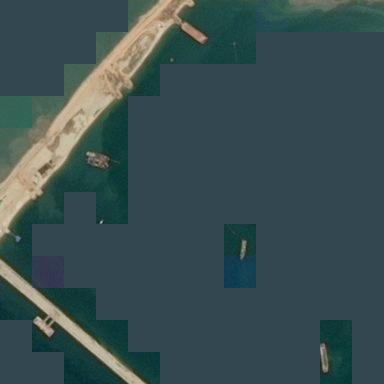}}
}{\small \(12\times 12\)}\hspace{.1cm}%
\resizebox{0.2\linewidth}{!}{\includegraphics{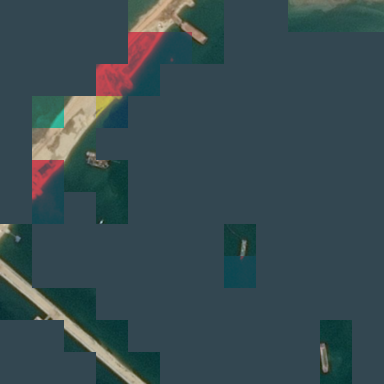}}\hspace{.175cm}
}

\end{subfigure}
\end{center}

  \vspace{-10pt} 
    \caption{
An examples of the instance-based selection for the dataset. Missing pixels were filled with average color based on the instances given in the training set (image size: $384 \times 384$ pixels).
    }
    \label{fig:dyn_example2_sup}
    \vspace{-3pt}
\end{figure*}

\subsection{Dynamic Selection Masks.} \label{exp:dynamic}
Finally, we conducted experiments for instance-based mask selections. 
For the sake of comparison, we also considered dynamic mask selection on \cifar, \fashmnist, and \mnist\ to illustrate the additional benefits of the instance-based selection over the other competitors mentioned above, see again Figure~\ref{fig:dyn_example}. 
The thumbnail model~\(\maskmodel\) used was a small linear layer. %
The \(\lossmask\) reported is the average over all instance-dependent losses per batch.
The static selection approach starts dropping in performance at a mask loss \(\lossmask\) of $0.125$, while the dynamic mask approach could retain the accuracy.
A similar effect can be observed on \fashmnist\ and \cifar, although the differences are much more apparent for \cifar.
Overall, since the dynamic approach yielded instance-based masks, less pixels were selected.
It is worth stressing that for the dynamic approach, one has to transfer the selection mask per instance. 
For pixel-wise selection, this might only pay off if a few pixels are selected (since the coordinates per selected pixel need to be transferred as well). 
This problem is alleviated by selecting blocks, as shown next.

The potential of the dynamic selection approach is demonstrated on the \ships\ dataset. 
Here, we considered thumbnails of size \(12 \times 12\) and \(48 \times 48 \), respectively, which are processed by the thumbnail model $\maskmodel$ to identify the relevant parts of the input data per instance, see again Figure~\ref{fig:dynmasks}.
We used a small convolutional network as thumbnail model $\maskmodel$.\footnote{Composed of one selective kernel convolution~\cite{li2019selective} with 2 convolutional layers, 8 filters per layer, kernel size 3, a dilation rate of 1 and 2, respectively, followed by point-wise convolution.}
In Figure~\ref{fig:dyn_example2}~(a), the difference between dynamic and static block selection is shown (\(12\times 12\)-block masks).
Static selection quickly becomes inferior for this dynamic task.
In contrast, both, the \(12 \times 12\) and the \(48 \times 48 \) thumbnail selection performed well, see Figure~\ref{fig:dyn_example2}~(b).
\begin{figure}[!t]
 \vspace{.1cm}
\begin{subfigure}[t]{.47\linewidth}
\centering
   \resizebox{\linewidth}{!}{
        \input{figures/experiments/dyn_stat_ships}
      }
    \caption{static vs. dynamic}
\end{subfigure}\hfill\begin{subfigure}[t]{.47\linewidth}
\centering
   \resizebox{\linewidth}{!}{
      \input{figures/experiments/dyn_ships.tex}
      }
\caption{\(12 \times 12\) vs. \(48 \times 48 \)}
\end{subfigure}
\vspace{-3pt}
\caption{
Comparing dynamic and static mask selection as well as different block sizes on the \ships~dataset.
\label{fig:dyn_example2}
}
\end{figure}
The \(\lossmask\) loss per block was \(\frac{1}{12 \times 12 \times \ichannel}\) and  \(\frac{1}{48 \times 48 \times \ichannel}\), respectively.
Overall, only a fraction of the original data needs to be transferred.
For instance, using the model with \(\lossmask \approx 0.025\) and $12 \times 12$ thumbnails, one would on average transfer less than $100 \cdot (0.025 + 2 \cdot \frac{144}{384 \cdot 384}) \approx 2.7\%$  of the original data per instance.
Figure~\ref{fig:dyn_example2_sup} shows two instances along with the (dynamic) masks for different mask loss \(\lossmask\).
The unimportant blocks (\eg, water or ground) were removed, but blocks containing ships or landmass remained.

\section{Conclusions.}
The transfer of data between servers and clients can become a major bottleneck during the inference phase of a neural network. 
We propose a framework that allows to automatically select the relevant parts of the input data needed by a model. 
Our approach resorts to various types of selection masks (\texttt{channel}, \texttt{pixel}, and \texttt{block}) that are optimized together with any given task model during the training phase. 
In addition to static selection, we also introduce instance-based selection to deal with tasks with shifting feature importance such as segmentation. 
Our experiments show that it is often possible to achieve a good accuracy with significantly less input data that needs to be transferred.

In addition, our mask selections are inferred from the task model. 
Hence, the selections eventually made also provides insights into what is considered relevant by the model and can, therefore, help to identify and understand learned patterns and potential biases.

\section*{Acknowledgement.}
We acknowledge support from the Independent Research Fund Denmark through the grant \emph{Monitoring Changes in Big Satellite Data via Massively-Parallel Artificial Intelligence} (9131-00110B).
We also acknowledge support by the Villum Foundation through the project \emph{Deep Learning and Remote Sensing for Unlocking Global Ecosystem Resource Dynamics (DeReEco)}.

\bibliographystyle{abbrv}
\bibliography{literature}

\begin{thebibliography}{10}

\bibitem{SDSS2019}
D.~Aguado, H.~J{\"o}nsson, H.~Zou, and {et al.}
\newblock The fifteenth data release of the sloan digital sky surveys: First
  release of manga-derived quantities, data visualization tools, and stellar
  library.
\newblock {\em Astrophysical Journal, Sup. Series}, 240(2), 2019.

\bibitem{aihara2011eighth}
H.~Aihara, C.~A. Prieto, D.~An, S.~F. Anderson, {\'E}.~Aubourg, E.~Balbinot,
  T.~C. Beers, A.~A. Berlind, S.~J. Bickerton, D.~Bizyaev, et~al.
\newblock The eighth data release of the sloan digital sky survey: first data
  from sdss-iii.
\newblock {\em The Astrophysical Journal Sup. Series}, 193(2):29, 2011.

\bibitem{pmlr-v97-balin19a}
M.~F. Bal{\i}n, A.~Abid, and J.~Zou.
\newblock Concrete autoencoders: Differentiable feature selection and
  reconstruction.
\newblock In {\em {ICML}}, pages 444--453. PMLR, 2019.

\bibitem{Breiman2001}
L.~Breiman.
\newblock Random forests.
\newblock {\em Machine Learning}, 45(1):5--32, 2001.

\bibitem{CoatesHWWCN13}
A.~Coates, B.~Huval, T.~Wang, D.~J. Wu, B.~C. Catanzaro, and A.~Y. Ng.
\newblock Deep learning with {COTS} {HPC} systems.
\newblock In {\em {ICML}}, volume~28, pages 1337--1345. JMLR.org, 2013.

\bibitem{discreteModelComp}
S.~Gao, F.~Huang, J.~Pei, and H.~Huang.
\newblock Discrete model compression with resource constraint for deep neural
  networks.
\newblock In {\em {CVPR}}, June 2020.

\bibitem{geurts2006extremely}
P.~Geurts, D.~Ernst, and L.~Wehenkel.
\newblock Extremely randomized trees.
\newblock {\em Machine learning}, 63(1):3--42, 2006.

\bibitem{GordonENCWYC18}
A.~Gordon, E.~Eban, O.~N. andcit Bo~Chen, H.~Wu, T.~Yang, and E.~Choi.
\newblock Morphnet: Fast {\&} simple resource-constrained structure learning of
  deep networks.
\newblock In {\em {CVPR}}, pages 1586--1595. {IEEE}, 2018.

\bibitem{SKA2017}
K.~Grange, B.~Alachkar, S.~Amy, D.~Barbosa, P.~Bommimmeni, and P.~Boven.
\newblock Square kilometre array: The radio telescope of the xxi century.
\newblock {\em Astronomy Reports}, 61(4):288--296, 2017.

\bibitem{han2018autoencoder}
K.~Han, Y.~Wang, C.~Zhang, C.~Li, and C.~Xu.
\newblock Autoencoder inspired unsupervised feature selection.
\newblock In {\em ICASSP}, pages 2941--2945. IEEE, 2018.

\bibitem{Han2015}
S.~Han, J.~Pool, J.~Tran, and W.~J. Dally.
\newblock Learning both weights and connections for efficient neural networks.
\newblock In {\em {NeurIPS}}, pages 1135--1143, Cambridge, MA, USA, 2015. MIT
  Press.

\bibitem{binNetworks}
I.~Hubara, M.~Courbariaux, D.~Soudry, R.~El-Yaniv, and Y.~Bengio.
\newblock Binarized neural networks.
\newblock In {\em {NeurIPS}}, volume~29. Curran Associates, 2016.

\bibitem{LSST2019}
{\v Z}.~{Ivezi{\'c}}, S.~M. {Kahn}, J.~A. {Tyson}, B.~{Abel}, E.~{Acosta},
  R.~{Allsman}, D.~{Alonso}, Y.~{AlSayyad}, S.~F. {Anderson}, J.~{Andrew}, and
  et~al.
\newblock {LSST: From Science Drivers to Reference Design and Anticipated Data
  Products}.
\newblock {\em The Astrophysical Journal}, 873:111, 2019.

\bibitem{adam}
D.~P. Kingma and J.~Ba.
\newblock Adam: {A} method for stochastic optimization.
\newblock {\em {CoRR}}, abs/1412.6980, 2014.

\bibitem{cifar10}
A.~Krizhevsky, G.~Hinton, et~al.
\newblock Learning multiple layers of features from tiny images.
\newblock 2009.

\bibitem{pmlr-v70-kumar17a}
A.~Kumar, S.~Goyal, and M.~Varma.
\newblock Resource-efficient machine learning in 2 {KB} {RAM} for the {IoT}.
\newblock In {\em {ICML}}, volume~70, pages 1935--1944. PMLR, 2017.

\bibitem{lecun2010mnist}
Y.~LeCun, C.~Cortes, and C.~Burges.
\newblock {MNIST} handwritten digit database.
\newblock {\em AT\&T Labs}, 2010.

\bibitem{lemhadri2021lassonet}
I.~Lemhadri, F.~Ruan, L.~Abraham, and R.~Tibshirani.
\newblock Lassonet: A neural network with feature sparsity.
\newblock {\em {JMLR}}, 22(127):1--29, 2021.

\bibitem{Sentinel}
J.~Li and D.~P. Roy.
\newblock A global analysis of sentinel-2a/2b and landsat-8 data revisit
  intervals and implications for ter. monitoring.
\newblock {\em Remote Sensing}, 9(902), 2017.

\bibitem{li2019selective}
X.~Li, W.~Wang, X.~Hu, and J.~Yang.
\newblock Selective kernel networks.
\newblock In {\em {CVPR}}, pages 510--519. IEEE, 2019.

\bibitem{lintott2008galaxy}
C.~J. Lintott, K.~Schawinski, A.~Slosar, K.~Land, S.~Bamford, D.~Thomas, M.~J.
  Raddick, R.~C. Nichol, A.~Szalay, D.~Andreescu, et~al.
\newblock Galaxy zoo: morphologies derived from visual inspection of galaxies
  from the sloan digital sky survey.
\newblock {\em Monthly Notices of the Royal Astronomical Society},
  389(3):1179--1189, 2008.

\bibitem{lu2007feature}
Y.~Lu, I.~Cohen, X.~S. Zhou, and Q.~Tian.
\newblock Feature selection using principal feature analysis.
\newblock In {\em ACM Multimedia}, pages 301--304, 2007.

\bibitem{lu2018deeppink}
Y.~Lu, Y.~Fan, J.~Lv, and W.~S. Noble.
\newblock Deeppink: reproducible feature selection in deep neural networks.
\newblock In {\em {NeurIPS}}, pages 8676--8686, 2018.

\bibitem{gumbelTrick2014}
C.~J. Maddison, D.~Tarlow, and T.~Minka.
\newblock A{*} sampling.
\newblock In {\em {NeurIPS}}, pages 3086--3094, 2014.

\bibitem{NIPS2016_6250}
F.~Nan, J.~Wang, and V.~Saligrama.
\newblock Pruning random forests for prediction on a budget.
\newblock In {\em {NeurIPS}}, pages 2334--2342. Curran Associates, 2016.

\bibitem{37648}
Y.~Netzer, T.~Wang, A.~Coates, A.~Bissacco, B.~Wu, and A.~Y. Ng.
\newblock Reading digits in natural images with unsupervised feature learning.
\newblock In {\em {NeurIPS} Workshop}, 2011.

\bibitem{PRISHCHEPOV2012195}
A.~V. Prishchepov, V.~C. Radeloff, M.~Dubinin, and C.~Alcantara.
\newblock The effect of landsat {ETM}/{ETM}+ image acquisition dates on the
  detection of agricultural land abandonment in eastern europe.
\newblock {\em Remote Sensing of Environment}, 126:195 -- 209, 2012.

\bibitem{amsgrad}
S.~J. Reddi, S.~Kale, and S.~Kumar.
\newblock On the convergence of adam and beyond.
\newblock In {\em {ICLR}}, 2018.

\bibitem{Reichstein19}
M.~Reichstein, G.~Camps-Valls, B.~Stevens, M.~Jung, J.~Denzler, N.~Carvalhais,
  and Prabhat.
\newblock Deep learning and process understanding for data-driven earth system
  science.
\newblock {\em Nature}, 566(7743):195--204, 2019.

\bibitem{lossyAE}
L.~Theis, W.~Shi, A.~Cunningham, and F.~Husz{\'{a}}r.
\newblock Lossy image compression with compressive autoencoders.
\newblock In {\em {ICLR}}. OpenReview.net, 2017.

\bibitem{Landsat}
M.~A. Wulder, J.~G. Masek, W.~B. Cohen, T.~R. Loveland, and C.~E. Woodcock.
\newblock Opening the archive: How free data has enabled the science and
  monitoring promise of landsat.
\newblock {\em Remote Sensing of Environment}, 122(Sup. C):2 -- 10, 2012.
\newblock Landsat Legacy Special Issue.

\bibitem{fmnist}
H.~Xiao, K.~Rasul, and R.~Vollgraf.
\newblock Fashion-mnist: a novel image dataset for benchmarking machine
  learning algorithms.
\newblock {\em {CoRR}}, cs.LG/1708.07747, 2017.

\bibitem{xu2018scaling}
X.~Xu, Y.~Ding, S.~X. Hu, M.~Niemier, J.~Cong, Y.~Hu, and Y.~Shi.
\newblock Scaling for edge inference of deep neural networks.
\newblock {\em Nature Electronics}, 1(4):216--222, 2018.

\bibitem{XuKWC2013}
Z.~E. Xu, M.~J. Kusner, K.~Q. Weinberger, and M.~Chen.
\newblock Cost-sensitive tree of classifiers.
\newblock In {\em {ICML}}, volume~28, pages 133--141. {JMLR.org}, 2013.

\bibitem{exploreModelComp}
Y.~Zhang, S.~Gao, and H.~Huang.
\newblock Exploration and estimation for model compression.
\newblock In {\em {ICCV}}, pages 487--496, 2021.

\end{thebibliography}

\appendix

\section{Reproducibility}\label{A:rep}

We provide all the source code used for our experiments via an open GitHub repository.\footnote{\url{https://github.com/StefOe/selection-masks}} %
For training the models and assessing their quality, different machines, and GPU devices (NVIDIA K20, K40, GTX1080, and V100) were used.

The chosen parameters for our static and dynamic experiments are summarized in Table~\ref{tab:static_parameters} and \ref{tab:dyn_parameters}, respectively. 
The scripts with our parameter settings to run the corresponding experiments can be found in the code repository (directory \texttt{exp\_scripts}).
The data transformations applied during training are listed in Table~\ref{tab:transforms}. 

\begin{table}[h]
    \centering
    \setlength\tabcolsep{5.1pt}
    \caption{Dataset augmentations and transforms}
    \begin{tabular}{lcccc}
\toprule
dataset & vert.\ flip & horiz.\ flip & crop & normalize\\
\midrule
\mnist & & & & \Checkmark \\
\fashmnist & & & & \Checkmark \\
\cifar & &  \Checkmark &  \Checkmark & \Checkmark \\
\svhn & &  \Checkmark &  \Checkmark & \Checkmark \\
\galaxy & \Checkmark & \Checkmark &  \Checkmark & \Checkmark \\
\remotedataset & \Checkmark & \Checkmark & \Checkmark & \Checkmark \\
\ships & \Checkmark & \Checkmark & \Checkmark & \Checkmark \\
\bottomrule
    \end{tabular}
    \label{tab:transforms}
\end{table}

To run the experiments, the necessary requirements need to be installed (\eg, via \texttt{pip install -r requirements.txt}).
The individual experiments can be started via the command line. 
For instance, the \texttt{block} experiment on \svhn\ (using the normal ``any'' selection) can be started via:
\begin{quote}
    \texttt{python create\_mask.py \\ -{}-dataset svhn ~-{}-mask-type static \\ ~-{}-any-granularity subCXLIVdrant \\ ~-{}-lambda-patience 5 ~-{}-lambda-init~0.125 \\ ~-{}-lambda-factor 1.25 -{}-n-epochs 300 \\ ~-{}-use-warmup-net 1 -{}-lr 0.001 \\ ~-{}-n-repeats 10}
\end{quote}

By executing this command, a corresponding log file and checkpoints in the directory \path{runs} will be created.
Note that the flag \texttt{-{}-use-warmup-net 1} enables the use of pre-trained models and optimizers.
The mask type (\ie, random, static, or dynamic) can be changed via the flag \texttt{-{}-mask-type}, which is, together with the flag \texttt{-{}-dataset}, the only required parameter.
For instance, the dynamic mask experiment for \mnist\ can be reproduced via the following command:
\begin{quote}
        \texttt{python create\_mask.py \\ ~-{}-dataset mnist -{}-mask-type dynamic \\ -{}-dynamic-mask linear \\ ~-{}-any-granularity subpixel \\ ~-{}-lambda-patience 5 \\ ~-{}-lambda-init 0.0005 \\ ~-{}-lambda-factor 1.5 -{}-n-epochs 400 \\ ~-{}-use-warmup 1 -{}-lr 0.0005 \\ ~-{}-n-repeats 10}
\end{quote}
The remaining experiments can be started in a similar fashion. 
An overview over the flags and options can be obtained via \texttt{python create\_mask.py -{}-help}.

\begin{table}[h]

    \centering
\begin{subtable}[h]{1\linewidth}
    \centering
    \setlength\tabcolsep{3.5pt}
    
    \caption{Parameters used for static experiments}
    \resizebox{1.0\linewidth}{!}{
    \begin{tabular}{llrrrrrr}
\toprule
granularity & dataset & \(\lambdainit\) & \(\lambdafactor\) & \(\patience\) & lr mask & epochs & any-init \\
\midrule
\texttt{channel} &  \remotedataset & 0.1 & 1.25 & 10 & 0.001 & 300 & 3 \\
\addlinespace
\texttt{pixel}& \cifar &  1 & 1.05 & 5 & 0.001 & 300 & 3 \\
&  \mnist &  0.0005 & 1.5 & 5 & 0.005 & 400 & 3 \\
&  \fashmnist &  0.0005 & 1.5 & 5 & 0.005 & 400 & 3 \\
&  \galaxy &  1 & 1.5 & 2 & 0.001 & 300 & 3 \\
\addlinespace
\texttt{block}\\
\quad(\(12\times 12\))  &  \svhn & 0.125  & 1.25 & 5 & 0.001 & 300 & 3 \\
\bottomrule
    \end{tabular}
    }
    \label{tab:static_parameters}
\end{subtable}
\vspace{.5cm}
    \centering
\begin{subtable}[h]{1\linewidth}
    \centering
    \setlength\tabcolsep{2.5pt}
    \caption{Parameters used for dynamic experiments}
    \resizebox{1.0\linewidth}{!}{
    \begin{tabular}{lllrrrrrr}
\toprule
granularity & mask model \(\maskmodel\) & dataset & \(\lambdainit\) & \(\lambdafactor\) & \(\patience\) & lr mask & epochs\\
\midrule
\texttt{pixel} & linear & \mnist & 0.0005 & 1.5 & 5 & 0.0005 & 400 \\
&  & \fashmnist & 0.0005 & 1.5 & 2 & 0.0005 & 400 \\
& convatt & \cifar & 0.1 & 1.15 & 10 & 0.001 & 300 \\
\addlinespace
\texttt{block} \\
\quad(\(12\times 12\)) & convatt & \ships & 0.025 & 1.15 & 9 & 0.0005 & 400 \\
\quad(\(48\times 48\)) & convatt & \ships & 0.025 & 1.15 & 9 & 0.0005 & 400 \\
\bottomrule
    \end{tabular}
    }
    \label{tab:dyn_parameters}
\end{subtable}
\end{table}

\section{Broader Impact}
We expect that such selection masks will play an important role for many data-intensive domains to alleviate the data transfer problem between centralized storage servers and clients:
In many cases, the collected data are stored on centralized storage servers. The transfer of data between such servers and the users has already become a severe bottleneck. 
For instance, practitioners in remote sensing already resort to reduced versions of the data since a full data transfer would be too time-consuming (\ie, instead of the full multi-spectral image data, reduced versions are often considered, such as channels computed via the so-called Normalized Difference Moisture Index (NDMI)).
A similar situation is given in astronomy, where the data transfer will become a key bottleneck in future with projects producing exabytes of data per year~\cite{SDSS2019,SKA2017,LSST2019}.
Today's storage servers already provide advanced APIs to select or crop the data prior to the transmission (\eg, the Planet API mentioned above can be used to select pieces of the data beforehand; similarly, services such as the Copernicus Open Access Hub allow to select subsets of the data and also provide previews of the data). 
The methods presented in this work offer the potential to alleviate the data transfer bottleneck in such domains, which, in the end, will lead to less resources that have to be spent for the overall infrastructure (\eg, more powerful servers, faster network connections, \ldots).
Naturally, while our work addresses the two aforementioned application domains, the approaches developed are generally applicable to other domains as well (\eg, IoT data, medical image data, \ldots). 
We therefore believe that the results presented in this work will affect a broad range of domains and applications.

\end{document}